\documentclass[sigconf, acmart]{acmart}

\AtBeginDocument{%
  \providecommand\BibTeX{{%
    \normalfont B\kern-0.5em{\scshape i\kern-0.25em b}\kern-0.8em\TeX}}}

\setcopyright{acmcopyright}
\copyrightyear{2022}
\acmYear{2022}
\setcopyright{acmlicensed}\acmConference[KDD '22]{Proceedings of the 28th ACM SIGKDD Conference on Knowledge Discovery and Data Mining}{August 14--18, 2022}{Washington, DC, USA}
\acmBooktitle{Proceedings of the 28th ACM SIGKDD Conference on Knowledge Discovery and Data Mining (KDD '22), August 14--18, 2022, Washington, DC, USA}
\acmPrice{15.00}
\acmDOI{10.1145/3534678.3539210}
\acmISBN{978-1-4503-9385-0/22/08}

\usepackage{booktabs} 
\usepackage{algorithmic}
\usepackage[linesnumbered, ruled, vlined]{algorithm2e}
\usepackage{multirow}
\usepackage{tablefootnote}
\usepackage{graphicx}
\usepackage{subfigure}
\usepackage{flushend}
\usepackage{color}
\usepackage{enumitem}
\usepackage{array, makecell}
\usepackage{float}
\usepackage{balance}

\usepackage{xcolor}
\usepackage{xspace}
\usepackage{arydshln}

\newcommand{\method}{OAG-BERT\xspace}

\usepackage[]{threeparttable}

\usepackage{placeins}

\newcommand{\vpara}[1]{\vspace{0.05in}\noindent\textbf{#1}\xspace\xspace}

\newcommand{\vcaption}[1]{\vspace{-0.02in}\caption{#1}\vspace{-0.12in}}

\newcommand{\hide}[1]{} 

\begin{document}


\title{\method: Towards A Unified Backbone Language Model For Academic Knowledge Services
}

\author{Xiao Liu}
\affiliation{Tsinghua University}
\email{liuxiao21@mails.tsinghua.edu.cn}
\authornote{The authors contributed equally to this research.}

\author{Da Yin}
\affiliation{Tsinghua University}
\email{yd18@mails.tsinghua.edu.cn}
\authornotemark[1]

\author{Jingnan Zheng}
\affiliation{National University of Singapore}
\email{e0718957@u.nus.edu}

\author{Xingjian Zhang}
\affiliation{Tsinghua University}
\email{zhangxj18@mails.tsinghua.edu.cn}

\author{Peng Zhang}
\affiliation{Zhipu AI}
\email{zpjumper@gmail.com}

\author{Hongxia Yang}
\affiliation{DAMO Academy, Alibaba Group}
\email{yang.yhx@alibaba-inc.com}

\author{Yuxiao Dong}
\affiliation{Tsinghua University}
\email{yuxiaod@tsinghua.edu.cn}

\author{Jie Tang}
\authornote{Jie Tang is the corresponding author.}
\affiliation{Tsinghua University}
\email{jietang@tsinghua.edu.cn}

\renewcommand{\shortauthors}{Xiao Liu et al.}
\begin{abstract}

Academic knowledge services have substantially facilitated the development of the science enterprise by providing a plenitude of efficient research tools. 
However, many applications highly depend on ad-hoc models 
and 
expensive human labeling to understand scientific contents, 
hindering deployments into real products. 
To build a unified backbone language model for different knowledge-intensive academic  applications, we pre-train an academic language model \method that integrates both the heterogeneous entity knowledge and scientific corpora in the Open Academic Graph (OAG)---the largest public academic graph to date.  
In \method, we develop strategies for pre-training text and entity data along with zero-shot inference techniques. 
\method achieves outperformance over baselines on nine academic tasks including two demo applications, demonstrating its potential to serve as one foundation model for academic knowledge services. 
Its zero-shot capability furthers the path to mitigate the need of expensive annotations. 
\method has been deployed for real-world applications, such as the reviewer recommendation function for National Nature Science Foundation of China (NSFC)---one of the largest funding agencies in China---and paper tagging in AMiner (\url{https://www.aminer.cn}). 
All codes and pre-trained models are available via the CogDL toolkit\footnote{\url{https://github.com/thudm/oag-bert}.}.

\hide{
To enrich language models with domain knowledge is crucial but difficult. Based on the world's largest public academic graph Open Academic Graph (OAG), we pre-train an academic language model, namely \method, which integrates massive heterogeneous entities including paper, author, concept, venue, and affiliation. To better endow \method with the ability to capture entity information, we develop novel pre-training strategies including heterogeneous entity type embedding, entity-aware 2D positional encoding, and span-aware entity masking. For zero-shot inference, we design a special decoding strategy to allow \method to generate entity names from scratch. We evaluate the \method on various downstream academic tasks, including NLP benchmarks, zero-shot entity inference, heterogeneous graph link prediction and author name disambiguation. Results demonstrate the effectiveness of the proposed pre-training approach to both comprehending academic texts and modeling knowledge from heterogeneous entities. \method has been deployed to multiple real-world applications, such as reviewer recommendations\hide{ for NSFC (National Nature Science Foundation of China)} and paper tagging in the AMiner system. \method\footnote{\url{https://github.com/thudm/oag-bert}} is also available to the public through the CogDL package.
}
\end{abstract}

%
%



\maketitle


\section{Introduction}

\hide{
\begin{figure}
    \centering
    \includegraphics[width=1\columnwidth,trim={0.1cm 0.6cm 0 0},clip]{figs/example.png}
    \caption{\method outperforms SciBERT on a range of entity-related tasks by 2.75\%-21.0\% (Absolute Gain).}
    \label{fig:example}
    \vspace{-0.2in}
\end{figure}
}

Academic knowledge services, 
such as AMiner~\cite{tang2008arnetminer}, 
Google Scholar, 
Microsoft Academic Service~\cite{wang2020microsoft}, 
and Semantic Scholar, 
have been of great assistance to advance the science enterprise. 
Beyond collecting statistics, e.g., citation count, an increasing attention of these platforms has been focused on providing AI-powered academic knowledge applications, including paper recommendation~\cite{du2019polar++,cohan2020specter}, expert matching~\cite{qian2018weakly}, taxonomy construction~\cite{shen2020taxoexpan}, and knowledge evolution~\cite{yin2021mrt}.

However, most of these applications 
are built with specified models 
to understand scientific contents. 
For example, OAG~\citeauthor{zhang2019oag} employs the doc2vec~\cite{le2014distributed} embeddings trained on a small corpus for academic entity alignment~\cite{zhang2019oag}. 
Zhang et al.~\cite{Zhang:MatchWWW21} leverage an attention strategy to model the text and metadata embeddings for paper tagging. 
In other words, the success of such academic systems heavily rely on different language understanding components. 
In addition, task-specific annotated datasets required by these components demand arduously expensive labeling cost.

The newly emerging pre-trained models,
such as BERT~\cite{devlin2018bert} and GPT~\cite{radford2019language}, have substantially promoted the development of natural language processing (NLP). 
Specifically, pre-trained language models for academic data have also been developed, such as BioBERT~\cite{lee2020biobert} for the biomedical field and SciBERT~\cite{beltagy2019scibert} for scientific literature. 
However, these models mainly focus on scientific texts and ignore the connected entity knowledge that can be crucial for many knowledge-intensive applications. 
For example, in author name disambiguation~\cite{zhang2018name,chen2020conna}, the affiliations of a paper's authors offer important signals about their identities. 

In light of these issues, we propose to pre-train a unified entity-augmented academic language model, \method, as the backbone model for diverse academic mining tasks and knowledge applications. 
\method is pre-trained from 
the Open Academic Graph (OAG)~\cite{zhang2019oag}, which is to date the largest publicly available heterogeneous academic entity graph. 
It contains more than 700 million entities (papers, authors, fields of study, venues, and affiliations), 2 billion relationships, and 5 million papers with full contents and 110 million abstracts as corpora. 


To handle the heterogeneous knowledge, we design the entity type embedding for each type of entities, respectively. 
To implement the masked language pre-training over entity names with various lengths, we leverage a span-aware entity masking strategy that can select to mask a continuous span of tokens according to the entity length. 
To better ``notify" the \method model with the entity span and sequence order, we propose the entity-aware 2D positional encoding to take both the inter-entity sequence order and intra-entity token order into consideration.

We apply  \method  to nine academic knowledge applications, including name disambiguation~\cite{zhang2018name,chen2020conna}, literature retrieval, entity graph completion~\cite{dong2017metapath2vec, hu2020heterogeneous}, paper recommendation~\cite{cohan2020specter}, user activity prediction~\cite{cohan2020specter}, fields-of-study tagging~\cite{marlow2006ht06}, venue prediction, affiliation prediction, and automatic title generation. 
Moreover, we present a number of prompt-based zero-shot usages of \method, including the predictions of a paper's venue, affiliations, and fields of study, in which the annotation cost is significantly mitigated. 

To sum up, we make the following contributions in this paper:

\begin{itemize}[leftmargin=0.15in]
    \item \textbf{A Unified Backbone Model \method:} We identify the challenge in existing academic knowledge applications, which heavily depend on ad-hoc models, corpora, and task-specific annotations. To address the problem, we present \method as a unified backbone model with 110M parameters to support it.
    
    \item \textbf{Entity-Augmented Language Model Pre-Training:} 
    In \method, we enrich the language model with the massive heterogeneous entity knowledge from OAG. 
    We design pre-training strategies to incorporate entity knowledge into the model. 
    
    \item \textbf{Prompt-based Zero-Shot Inference:} We design a decoding strategy to allow \method to perform well on prompt-based zero-shot inference, which offers the potential to significantly reduce the annotation cost in many downstream applications.
    
    \item \textbf{System Deployment and Open \method Model:} We demonstrate the effectiveness of \method on nine  academic knowledge applications. 
    In addition, \method has been deployed as the infrastructure of AMiner\footnote{\url{https://www.aminer.cn/}} and also used for NSFC's grant reviewer recommendation. 
    The pre-trained model  is open to public access through the CogDL~\cite{yukuo2021cogdl} package for free.
\end{itemize}

\section{Related Works}

The proposed \method model is based on BERT~\cite{devlin2018bert}, a self-supervised~\cite{liu2020self} bidirectional language model. It employs multi-layer transformers as its encoder and uses masked token prediction as its objective, allowing massive unlabeled text data as a training corpus. 
BERT has many variants. SpanBERT~\cite{joshi2020spanbert} develops span-level masking which benefits span selection tasks. ERNIE~\cite{zhang2019ernie} introduces explicit knowledge graph inputs to the BERT encoder and achieves significant improvements over knowledge-driven tasks.

As for the academic domain, previous works such as BioBERT~\cite{lee2020biobert} or SciBERT~\cite{beltagy2019scibert} leverage the pre-training process on scientific domain corpus and achieve state-of-the-art performance on several academic NLP tasks. The S2ORC-BERT~\cite{lo2019s2orc}, applies the same method with SciBERT on a larger scientific corpus and slightly improves the performance on downstream tasks. Later works~\cite{gururangan2020don} further show that continuous training on specific domain corpus also benefits the downstream tasks. These academic pre-training models rely on large scientific corpora. SciBERT uses the semantic scholar corpus~\cite{ammar2018construction}. Other large academic corpora including AMiner~\cite{tang2008arnetminer}, OAG~\cite{tang2008arnetminer,zhang2019oag}, and Microsoft Academic Graph (MAG)~\cite{kanakia2019scalable} also integrate massive publications with rich graph information as well, such as authors and research fields.

On academic graphs, some tasks involve not only text information from papers but also structural knowledge lying behind graph links. For example, to disambiguate authors or concepts with the same names~\cite{zhang2018name,chen2020conna,liu2021oag_know}, the model needs to learn node representations in the heterogeneous graph. To better recommend papers for online academic search~\cite{du2018polar,du2019polar++}, graph information including related academic concepts and published venues could provide great benefits. To infer experts' trajectory across the world~\cite{wu2018have}, associating authors with their affiliation on semantic level would help. Capturing features from paper titles or abstracts is far from enough for these types of challenges.


\section{\method: A LANGUAGE MODEL WITH ACADEMIC KNOWLEDGE}
\label{sec:methods}

We present the \method model with the goal of pre-training on both the academic corpus and heterogeneous entity graph. 
Overall, \method is a bidirectional Transformer-based ~\cite{vaswani2017attention,devlin2018bert} pre-training model with 12 Transformer encoder layers. 
We introduce improvements to the model architecture and the pre-training process to handle not only text data but also the entity graph.
\method is trained on the Open Academic Graph (OAG) ~\cite{zhang2019oag}---the largest public heterogeneous entity graph to date.

\begin{figure}[t]
    \centering
    \includegraphics[width=\linewidth]{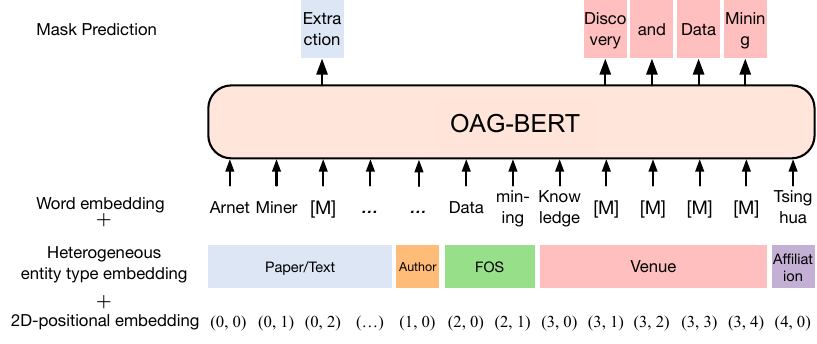}
    \caption{Heterogeneous entity augmentation in \method. \textmd{1) \textit{Heterogeneous entity type embedding} allows \method be aware of different types of entities, 2) \textit{Span-aware entity masking} selects a continuous span within long entities (such as the ``Knowledge Discovery and Data Mining''), and 3) \textit{Entity-aware 2D-positional embedding} jointly models inter and intra-entity token orders.}}
    \label{fig:framework}
    \vspace{-5mm}
\end{figure}


\subsection{The Model Architecture}\label{sec:strategies}

The key challenge of building \method 
is 
how to integrate the heterogeneous entity knowledge into language models. 
Most previous approaches~\cite{zhang2019ernie,liu2020k} focus on injecting homogeneous entities and relations from knowledge graphs like Wikidata, while heterogeneous entities for language models have been largely unexplored.


To augment the language model with the heterogeneous entity knowledge, we place the text of each paper in OAG (its title and abstract) and associated entities (its authors, their affiliations, fields of study, and venue) into a single sequence as one training instance. 
To further help \method model different types of entities, we leverage the following three techniques: \textit{heterogeneous entity type embedding}, \textit{entity-aware 2D-positional encoding}, and \textit{span-aware entity masking}.

\vpara{Heterogeneous Entity Type Embedding. }
In order to distinguish different types of entities, we propose to leverage the entity type embedding in the pre-training process to indicate the entity type, which is similar to the token type embedding used in BERT.

Figure~\ref{fig:framework} illustrates one paper example as the input of \method. 
The entities from the ``ArnetMiner'' paper are concatenated as an input sequence, whose length is limited to at most 512 tokens.
For each type of entities, including the paper title and abstract as the text entity, we label the text string with the original entity type index, e.g., 0 for text, 1 for authors, and so on. 
Additionally, different entities should be order-invariant in the input sequence. 
We thus shuffle their order in the sequence to let the model avoid learning any positional biases of these entities.



\vpara{Entity-Aware 2D-Positional Encoding.}
It is known that the Transformer model~\cite{vaswani2017attention} is permutation-invariant (i.e., unaware of the sequence order) and the critical technique to indicate the sequence order in natural language is to add \textit{positional embeddings}. 

However, the existing positional embeddings for texts~\cite{vaswani2017attention} are generally not applicable to capture the entity knowledge, as they cannot distinguish words from entities adjacent to each other and of the same type. 
For instance, if two affiliations ``Tsinghua University'' and ``University of California'' are  placed next to each other in a sequence, the vanilla Transformer model would assume that there is an affiliation named ``Tsinghua University University of California''.
Essentially, the positional embedding should be able to 1) imply the \textit{inter-entity} sequence order for distinguishing different entities, and 2)  indicate the \textit{intra-entity} token sequence order that is used as the traditional positional embedding.

We design the entity-aware 2D-positional embedding that solves both the inter-entity and intra-entity issues. 
Its first dimension is used for the inter-entity order, indicating which entity the token is in,  and its second dimension is for the intra-entity order, indicating the sequence of tokens. For a given position, the final positional embedding is calculated by adding the two positional embeddings together.

\vpara{Span-Aware Entity Masking.}
When performing masking, we adopt the same random masking strategy as in BERT for text contents such as paper titles and abstracts. 
But for the heterogeneous academic entities, we expect \method to memorize them well and thus develop a span-aware entity masking strategy that combines the advantages of both ERNIE~\cite{zhang2019ernie} and SpanBERT~\cite{joshi2020spanbert}. 

The intuition is that some of the entities are too long for \method to learn when using random masking at the single-token granularity. 
The span-aware entity masking strategy not only alleviates the problem, but also preserves the sequential relationship of an entity's tokens. 
Specifically, for an entity that has fewer than 4 tokens, we mask the whole entity; and for others, the masking length is decided by sampling from a geometric distribution $\mathrm{Geo}(p)$ that satisfies:
\begin{equation}
    p = 0.2~\rm{, and}~4\leq\mathrm{Geo(p)}\leq 10
\end{equation}
\noindent 
If the sampled length is less than the entity length, we only mask out the entity. 
For both text contents and entity contents, we mask 15\% of the tokens.

\vpara{Pre-LN BERT.}
In addition, we also adopt the Pre-LN BERT as used in DeepSpeed~\cite{rasley2020deepspeed}, where the layer normalization is placed inside the residual connection instead of after the add-operation in Transformer blocks. 
Previous work~\cite{zhang2020accelerating} demonstrates that training with Pre-LN BERT avoids vanishing gradients when using aggressive learning rates, making it more stable than the traditional Post-LN version for optimization.


\subsection{The Pre-Training Details}

The pre-training of \method is separated into two stages. 
In the first stage, we only use scientific texts (paper title, abstract, and body) as the model inputs, without using the entity augmented inputs introduced above. 
This process is similar to the pre-training of the original BERT model. 
We name the intermediate pre-trained model as the vanilla version of \method. 
In the second stage, based on the vanilla \method, we continue to train the model on the heterogeneous entities, including titles, abstracts, venues, authors, affiliations, and FOS.

\vpara{First Stage: Pre-train the vanilla \method.}
In the first stage of pre-training, we construct the training corpus from two sources: one comes from the PDF storage of AMiner;
and the other comes from the PubMed XML dump. We clean up and sentencize the corpus with SciSpacy~\cite{neumann2019scispacy}. The corpus adds up to around 5 million unique paper full-text from multiple disciplines. In terms of vocabulary, we construct our \method vocabulary using WordPiece, which is also used in the original BERT implementation. This ends up with 44,000 unique tokens in our vocabulary. The original BERT employs a sentence-level loss, namely Next Sentence Prediction (NSP), to learn the entailment between sequences, which has been found not that useful~\cite{joshi2020spanbert,liu2019roberta} and thus we do not adopt it. 

To better handle the entity knowledge of authors in the OAG, we transform the author name list into a sentence for each paper and place it between the title and abstract in the data prepossessing. Therefore, compared to previous models like SciBERT, our vocabulary contains more tokens from author names.
Following the training procedures of BERT, the vanilla \method is first pre-trained on samples with a maximum of 128 tokens and then shift to pre-training it over samples with 512 tokens.


\vpara{Second Stage: Enrich \method with entity knowledge.}
In the second stage of pre-training, we use papers and related entities from the OAG corpus. Compared to the corpus used in the first stage, we do not have full texts for all papers in OAG. Thus, we only use paper title and abstract as the paper text information. From this corpus, we select all authors with at least 3 papers published. Then we filter out all papers not linked to these selected authors. Finally, we got 120 million papers, 10 million authors, 670 thousand FOS, 53 thousand venues, and 26 thousand affiliations. Each paper and its connected entities are concatenated into a single training instance, following the input construction method described above. In this stage, we integrate the three strategies mentioned in Section \ref{sec:strategies} to endow \method the ability to ``notice'' the entities, rather than regarding them as pure texts. For tasks that require document-level representations, we present a version of \method with additional task-agnostic triplet contrast pre-training, which uses papers from the same authors in OAG as the positive pair and papers from authors with similar names as the negative pair.

\vspace{0.12in}
Our pre-training is conducted with 32 Nvidia Tesla V100 GPUs and an accumulated batch size of 32768. We use the default BERT pre-training configurations in deepspeed. We run 16K steps for the first stage pre-training and another 4K steps for the second stage.

\section{Applications}

We 
choose 9 fundamental academic mining tasks that are either directly deployed in the AMiner system or serve as prerequisites to other academic knowledge services, 
in which the entity knowledge may play an indispensable role. These applications feature 5 typical downstream applications:

\begin{itemize}
    \item Author Name Disambiguation~\cite{tang2011unified,zhang2018name,chen2020conna}
    \item Scientific Literature Retrieval~\cite{cohan2020specter,white2009scientific}
    \item Paper Recommendation~\cite{sugiyama2010scholarly,du2018polar,du2019polar++}
    \item User Activity Prediction~\cite{ye2013s,zhang2021understanding,cohan2020specter}
    \item Entity Graph Completion~\cite{dong2012link,hu2020heterogeneous,benchettara2010supervised}
\end{itemize}

\noindent and 4 prompt-based zero-shot applications without need for any annotations:

\begin{itemize}
    \item Fields-of-study Tagging~\cite{marlow2006ht06,shen2018web}
    \item Venue Prediction~\cite{yang2012venue,alhoori2017recommendation}
    \item Affiliation Prediction~\cite{wu2018careermap,wu2018have}
    \item Automatic Title Generation~\cite{qazvinian2008scientific,zhang2019multi}
\end{itemize}

We take SciBERT~\cite{beltagy2019scibert}
as our major compared method to demonstrate the importance of our entity-augmented pre-training. Other baselines compared are introduced individually in each section.

\subsection{Downstream Applications}

Compared to language models for common NLP tasks, backbone language models for academic mining are usually combined with other downstream supervised learning algorithms, such as clustering for name disambiguation~\cite{chen2020conna,zhang2018name} and graph neural networks for relation completion~\cite{hu2020heterogeneous}. This requires our language model to provide more informative representations on different entities.

Conventionally, these applications rely on individually trained representation upon their own small dataset or corpus. For example, in~\cite{hu2020heterogeneous}, to acquire embeddings for heterogeneous entities such as venues, fields-of-study and affiliations, the authors leverage metapath2vec~\cite{dong2017metapath2vec} embeddings which contains no semantic information; in~\cite{zhang2018name} authors use word2vec embeddings trained on a small portion of paper abstracts from AMiner systems. As an effort to unifying infrastructure for these applications, in the following evaluation \method reports to achieve better performance on all of them.

\begin{table}[H]
    \centering
    \vspace{-1mm}
    \caption{The Macro Pairewise F1 scores for the author name disambiguation competition whoiswho-v1.}
    \vspace{-3mm}
    \label{tab:name-disambiguation}
    \begin{threeparttable}
    \small
    \begin{tabular}{clrr}
    \toprule[1.2pt]
    & Inputs                      & SciBERT & \method \\
    \midrule
    \multirow{4}{*}{Unsupervised}   & \textit{title}              & 0.3690 & \textbf{0.4120} \\
                                    & \quad \textit{+fos}         & 0.4101 & \textbf{0.4643} \\
                                    & \quad \textit{+venue}       & 0.3603 & \textbf{0.4247} \\
                                    & \quad \textit{+fos+venue}   & 0.3903 & \textbf{0.4823} \\ \midrule
    \multirow{1}{*}{Supervised}     & Leader Board Top1           & \multicolumn{2}{c}{0.4900} \\
    \bottomrule[1.2pt]
    \end{tabular}
    \end{threeparttable}
    \vspace{-1mm}
\end{table}

\vpara{Author Name Disambiguation.}
Name disambiguation, or namely ``disambiguating who is who'', is a fundamental challenge for curating academic publication and author information, as duplicated names widely exist in our lives. For example, Microsoft Academic reports more than 10,000 authors named ``James Smith'' in United States~\cite{sinha2015overview}.
Without effective author name disambiguation algorithms, it is difficult to identify the belonging-ship of certain papers for supporting applications such as expert matching, citation counting and h-index computing.

Given a set of papers with authors of the same name, the problem is usually formulated as designing algorithm to separate these papers into clusters, where papers in the same cluster belong to the same author and different clusters represent different authors. We use the public dataset \textit{whoiswho-v1}~\cite{chen2020conna,zhang2018name}\footnote{https://www.aminer.cn/whoiswho} and apply the embeddings generated by pre-trained models to solve name disambiguation from scratch. 
Following dataset setting, for each paper, we use the paper title and other attributes such as FOS or venue as input. We average over all the output token embeddings for title as the paper embedding. Then, we build a graph with all papers as the graph nodes and set a threshold to select edges. The edges are between papers where the pairwise cosine similarity of their embeddings is larger than the threshold. Finally, for each connected component in the graph, we treat it as a cluster. We searched the thresholds from 0.65 to 0.95 on the validation set
and calculated the macro pairwise f1 score on test.

%


The results in Table~\ref{tab:name-disambiguation} indicate that the embedding of \method is significantly better than the SciBERT embedding while directly used in the author name disambiguation. We also observe that for SciBERT the best threshold is always 0.8 while this value for \method is 0.9, which reflects that the paper embeddings produced by \method are generally closer than the ones produced by SciBERT.

In Table~\ref{tab:name-disambiguation} we list a range of experimental results given title, field-of-study, and venue as inputs respectively. Though we attempted to use the abstract, author, and affiliation information, there is no performance improvement as expected. We speculate it is because these types of information are more complex to use, which might require additional classifier head or fine-tuning, as the supervised classification task mentioned above. In addition, we also report the top 1 score in the name disambiguation challenge leaderboard\footnote{\url{https://www.biendata.xyz/competition/aminer2019/leaderboard/}} and find that our proposed \method reaches close performance as compared with the top-1 ad-hoc model for the contest.


\begin{table}[H]
    \centering
    \caption{Scientific Literature Retrieval evaluation on OAG-QA (Top-100) between SciBERT and \method.}
    \vspace{-4mm}
    \small
    \renewcommand\tabcolsep{10pt}
    \renewcommand\arraystretch{0.9}
    \begin{tabular}{@{}lcc@{}}
\toprule[1.2pt]
                    & SciBERT      & \method      \\ \midrule
Geometry            & 0.097        & \bf 0.147        \\
Math. \& Stats.     & 0.099        & \bf 0.166        \\
Algebra             & \bf 0.071        & 0.069        \\
Calculus            & 0.091        & \bf 0.160         \\
Number theory       & 0.067        & \bf 0.085        \\
Linear algebra      & 0.111        & \bf 0.160         \\
Astrophysics        & 0.041        & \bf 0.072        \\
Quantum mechanics   & 0.047        & \bf 0.080         \\
Classical mechanics & 0.085        & \bf 0.197        \\
Chemistry           & 0.181        & \bf 0.216        \\
Biochemistry        & 0.146        & \bf 0.319        \\
Health care         & 0.041        & \bf 0.262        \\
Natural science     & 0.101        & \bf 0.277        \\
Algorithm           & 0.084        & \bf 0.209        \\
Neuroscience        & 0.054        & \bf 0.120         \\
Computer vision     & 0.035        & \bf 0.205        \\
Data mining         & 0.082        & \bf 0.161        \\
Deep learning       & 0.044        & \bf 0.138        \\
Machine learning    & 0.085        & \bf 0.177        \\
NLP                 & 0.05         & \bf 0.160        \\
Economics           & 0.055        & \bf 0.151        \\ \midrule
Average             & 0.079        & \bf 0.168        \\ \bottomrule[1.2pt]
\end{tabular}
    \label{tab:literature_retrieval}
    \vspace{-3mm}
\end{table}

\vpara{Scientific Literature Retrieval.}
Scientific literature retrieval, which assists researchers finding relevant scientific literature given their natural language queries, is closely related to a wide-range of top-level applications including publication search, citation prediction and scientific question and answering. For example, for a professional question like \textit{``Does sleeping fewer hours than needed cause common cold?''}, we may retrieve a related paper \textit{``Sick and tired: does sleep have a vital role in the immune system?''}.

We evaluate \method with triplet contrastive training over a fine-grained topic-specific literature retrieval dataset OAG-QA~\citep{tam2022parameter}, which is constructed by collecting high-quality pairs of questions and cited papers in answers from Online Question-and-Answers (Q\&A) forums (Stack Exchange and Zhihu). It consists of 22,659 unique query-paper pairs from 21 scientific disciplines and 87 fine-grained topics. Given each topic is accompanied by 10,000 candidate papers including the groundtruth, and their titles and abstracts are taken as the corpus. We compute the cosine similarity between output embeddings of the query and paper for ranking. Results in Table~\ref{tab:literature_retrieval} suggest that \method has a consistently better performance than SciBERT across 20 scientific disciplines .

\begin{table}[t]
\centering
\caption{Paper recommendation and User Activity Prediction (Co-View and Co-Read) on Scidocs~\cite{cohan2020specter}.}
\vspace{-2mm}
\renewcommand\tabcolsep{4pt}
\renewcommand\arraystretch{0.8}
\small
\begin{tabular}{lcccccc}
    \toprule[1.2pt]
    \multirow{2}{*}{Models}     & \multicolumn{2}{c}{Paper Rec.} & \multicolumn{2}{c}{Co-View} & \multicolumn{2}{c}{Co-Read} \\ \cmidrule(l){2-3} \cmidrule(l){4-5} \cmidrule(l){6-7}
                                & nDCG & P@1 & MAP & nDCG & MAP & nDCG \\ \midrule
    Random    & 51.3 & 16.8 & 25.2 & 51.6 & 25.6 & 51.9 \\
    doc2vec   & 51.7 & 16.9 & 67.8 & 82.9 & 64.9 & 81.6 \\
    Sent-BERT & 51.6 & 17.1 & 68.2 & 83.3 & 64.8 & 81.3\\
    SciBERT   & 52.1 & 17.9 & 50.7 & 73.1 & 47.7 & 71.1\\
    \method   & \bf 52.6 & \bf 18.6 & \bf 74.7 & \bf 86.3 & \bf 71.4 & \bf 84.7 \\ \bottomrule[1.2pt]
\end{tabular}
\vspace{-3mm}
\label{tab:paper_recommendation}
\end{table}

\vpara{Paper Recommendation \& User Activity Prediction.}
As the number of scientific publications keeps soaring up, paper recommendation is playing an increasingly crucial role in many online academic systems, and therefore it is important to evaluate a backbone model's ability in boosting a production recommendation system. 
We consider the situation when users are browsing certain papers in our systems, and we want to 1) recommend them related papers of the ones they are reading, 2) predict papers they simultaneously viewed (Co-View) or pdf-accessed (i.e., Co-Read) in a user's browser session. In practice, for paper recommendation, it is often conducted in an ensemble manner: together with cosine similarities of textual embeddings encoded by language models, we jointly take other features such as citation overlaps, clicking counts and author similarities into consideration, and train a classifier to make the final decision; for user activity prediction, we mainly measure co-viewed or co-read papers' textual similarities.

We adopt Scidocs~\cite{cohan2020specter} paper recommendation and user activity prediction dataset for offline evaluation, which is constructed from real user
clickthroughs and loggings in a publication search engine. The recommendation dataset consists of 22k samples, in which 20k clickings are used for training the recommender, 1k for validation, and 1k for testing. For Co-View and Co-Read dataset, each of them contains 30k papers. Besides SciBERT, we compare with common passage representation methods, including doc2vec~\cite{le2014distributed} and Sent-BERT~\cite{reimers2019sentence}. Results in Table~\ref{tab:paper_recommendation} show that \method with triplet contrastive training can bring a consistent gain over compared methods in both paper recommendation and user activity prediction setting.


\begin{figure*}[t]
  \centering
  \includegraphics[width=17.5cm]{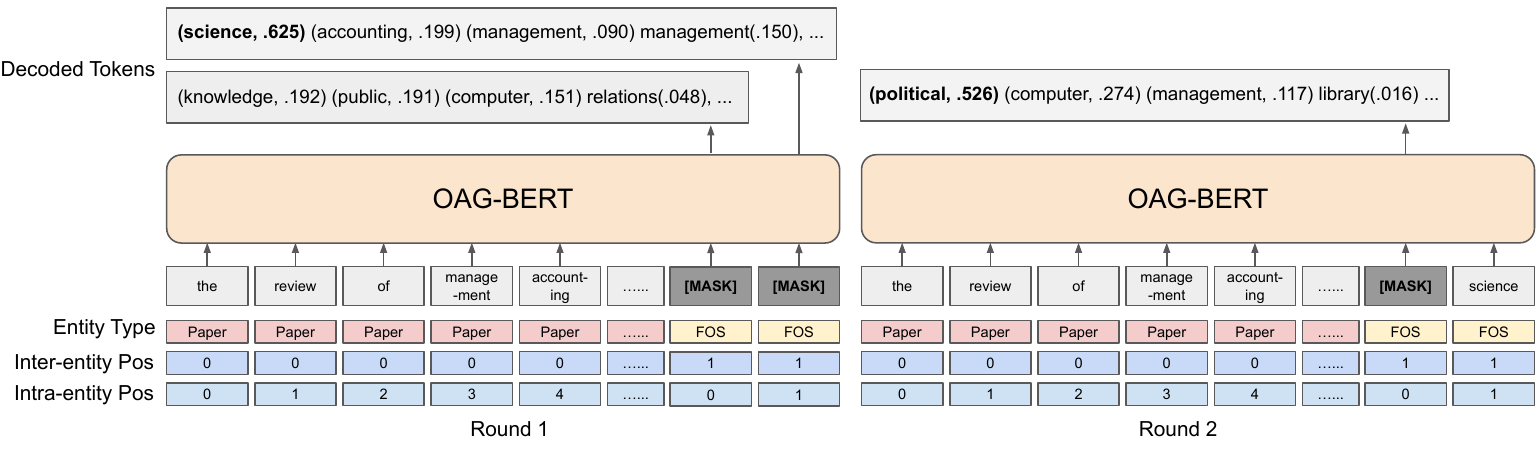}
  \vcaption{The decoding process of \method. \textmd{The left figure indicates that \method decodes the masked token ``science'' at the second position with the highest probability (0.625) for the first round. Then it decodes ``political'' at the first position with the highest probability (0.526) for the second round as shown in the right figure.}}
  \label{fig:decode-process}
\end{figure*}

\vpara{Entity Graph Completion.}
Academic entity graph, which consists of heterogeneous entities including papers, authors, fields-of-study, venues, affiliations and other potential entities with attributes, is a powerful organization form of academic knowledge and finds wide adoptions in many academic systems such as Microsoft Academic Graph~\cite{sinha2015overview} and AMiner~\cite{tang2008arnetminer}. However, such entity graphs have been suffered from the long-standing challenge of incomplete and missing relations, and therefore the task of entity graph completion becomes vital to their maintenance.

\begin{table}[t]
    \centering
    \vcaption{Results on Entity Graph Completion using HGT. \textmd{\method yields better initialization for heterogeneous entities.}}
    \label{tab:hgt}
    \renewcommand\tabcolsep{8pt}
    \small
    \begin{tabular}{lcccc}
    \toprule[1.2pt]
    \multirow{2}{*}{Models} & \multicolumn{2}{c}{Paper-Field} & \multicolumn{2}{c}{Paper-Venue} \\ \cmidrule(l){2-3} \cmidrule(l){4-5}
    & NDCG & MRR & NDCG & MRR \\
    \midrule
    XLNet & 0.3939 & 0.4473 & 0.4385 & 0.2584 \\
    SciBERT & 0.4740 & 0.5743 & 0.4570 & 0.2834 \\
    \method & \textbf{0.4892} & \textbf{0.6099} & \textbf{0.4844} & \textbf{0.3131} \\
    \bottomrule[1.2pt]
    \vspace{-0.2in}
    \end{tabular}
\end{table}

In this section, we apply the heterogeneous entity embeddings of \method as pre-trained initialization for entity embeddings on the academic graph and show that \method can also work together with other types of models. Specifically, we take the heterogeneous graph transformer (HGT) model from~\cite{hu2020heterogeneous}, a state-of-the-art graph neural network, to conduct entity graph completion pre-trained embeddings from \method.

To make predictions for the links in the heterogeneous graph, the authors of HGT first extract node features and then apply HGT layers to encode graph features. For paper nodes, the authors use XLNet~\cite{yang2019xlnet}, a well-known general-domain pre-trained language model, to encode titles as input features. For other types of nodes, HGT use metapath2vec~\cite{dong2017metapath2vec} to initialize the features.
However, XLNet was pre-trained on universal language corpus, lacking academic domain knowledge, and can only encode paper nodes by using their titles and is unable to generate informative embeddings for other types of nodes.

To this end, we replace the original XLNet encoder with our \method model, which can tackle the two challenges mentioned above. We use the \method model to encode all types of nodes and use the generated embeddings as their node features. To demonstrate the effectiveness of \method on encoding heterogeneous nodes, we also compare the performance of SciBERT with \method. We experimented on the CS dataset released by HGT\footnote{https://github.com/acbull/pyHGT}. The details of the dataset are delivered in the appendix.
The NDCG and MRR scores for the Paper-Field and Paper-Venue link prediction are reported in Table~\ref{tab:hgt}. It shows that SciBERT surpasses the original XLNet performance significantly, due to the pre-training on the large scientific corpus. Our proposed \method made further improvements on top of that, as it can better understand the entity knowledge on the heterogeneous graph.


\subsection{Prompt-based Zero-shot Applications}

\begin{table}[t]
  \centering
  \small
  \vcaption{The results for zero-shot inference tasks.}
  \label{tab:zero-shot-result}
  \begin{tabular}{lrrrrrr}
    \toprule[1.2pt]
    \multirow{2}{*}{Method} & \multicolumn{2}{c}{Paper Tagging} & \multicolumn{2}{c}{Venue} & \multicolumn{2}{c}{Affiliation} \\
    & Hit@1 & MRR & Hit@1 & MRR & Hit@1 & MRR \\
    \midrule
    SciBERT & 19.93\% & 0.37 & 9.87\% & 0.22 & 6.93\% & 0.19 \\
    \quad \textit{+prompt} & 29.59\% & 0.47 & 10.03\% & 0.21 & 8.00\% & 0.20 \\
    \quad \textit{+abstract} & 25.66\% & 0.43 & 18.00\% & 0.32 & 10.33\% & 0.22 \\
    \quad \textit{+both} & 35.33\% & 0.52 & 9.83\% & 0.22 & 12.40\% & 0.25\\
    \midrule
    \method & 34.36\% & 0.51 & 21.00\% & 0.37 & 11.03\% & 0.24 \\
    \quad \textit{+prompt} & 37.33\% & 0.55 & 22.67\% & 0.39 & 11.77\% & 0.25 \\
    \quad \textit{+abstract} & \textbf{49.59\%} & \textbf{0.67} & \textbf{39.00\%} & \textbf{0.57} & \textbf{21.67\%} & \textbf{0.38} \\
    \quad \textit{+both} & 49.51\% & \textbf{0.67} & 38.47\% & \textbf{0.57} & 21.53\% & \textbf{0.38} \\
    \bottomrule[1.2pt]
  \end{tabular}
  \vspace{-0.1in}
\end{table}

Despite \method's qualification in providing unified support to various downstream applications to get rid of ad-hoc models and corpus, a more challenging topic is to reduce task-specific annotations, which can be expensive in business deployment. 

Take affiliation prediction as an example, a common approach is to train a $k$-class classifier for $k$ given candidate institutions. However, as the progress of science, new universities, laboratories and companies emerge and to incorporate them into the pool may require re-annotating and re-training of the classifier with considerably high cost.

In light of the recent prompt-based~\cite{liu2021gpt} zero-shot and few-shot advances of large-scale pre-trained language models such as GPT-3~\cite{brown2020language}, in this section we also explore the potential of applying \method to zero-shot applications in academic mining. We discover that \method works surprisingly well on some fundamental applications, such as paper tagging, venue/affiliation prediction, and generation tasks such as title generation. We will first introduce how we implement the zero-shot inference on \method, and then the details of our applications.

\vpara{\method's zero-shot inference strategies.}
Although not using a unidirectional decoder structure like GPT-3, we find that the bidirectional encoder-based \method is also capable of decoding entities based on the knowledge it learned during pre-training. A running-example is provided in Figure~\ref{fig:decode-process}. In MLM, the token prediction can be seen as maximizing the probability of masked input tokens, treating predictions on each token independently by maximizing
$
  \sum_{w\in \textit{masked}}\log P(w|\mathcal{C})
$, 
where \textit{masked} is the collection of masked tokens and $\mathcal{C}$ denotes contexts. But in entity decoding, we cannot ignore the dependencies between tokens in each entity, and thus need to jointly consider the probability of all tokens in one entity as following
$
  \log P(w_1,w_2,...,w_l|\mathcal{C})
$, 
where $l$ is the entity length and $w_i$ is the $i$-th token in the entity. As MLM is not unidirectional model, the decoding order for the tokens in one entity can be arbitrary. Suppose the decoding order is $w_{i_1},w_{i_2},...,w_{i_l}$, where $i_1,i_2,...,i_l$ is a permutation of $1,2,...,l$. Then the prediction target can be reformed as maximizing
\begin{equation}
  \label{eq:entity-prob-decode}
  \sum_{1\le k\le l}\log P(w_{i_k}|\mathcal{C},w_{i_1},w_{i_2},...,w_{i_{k-1}})
\end{equation}
As the solution space is getting larger as $l$ increases, we adopt the following two strategies to determine the decoding order:

\begin{itemize}[leftmargin=0.15in]
    \item \textbf{Greedy:} we use greedy selection to decide the decoding order, by choosing the token with maximal probability to decode. An example is depicted in Figure~\ref{fig:decode-process}.
    \item \textbf{Beam search:} we can also use beam search~\cite{Tillmann2003WordRA} to search the token combinations with the highest probability.
\end{itemize}

Another challenge lies in choosing the appropriate entity length. Instead of using a fixed length, we traverse all entity lengths in a pre-defined range depending on the entity type and choose top candidates according to the calculated probability in Equation~\ref{eq:entity-prob-decode}.

\begin{figure}[t]
    \centering
    \includegraphics[width=1\linewidth]{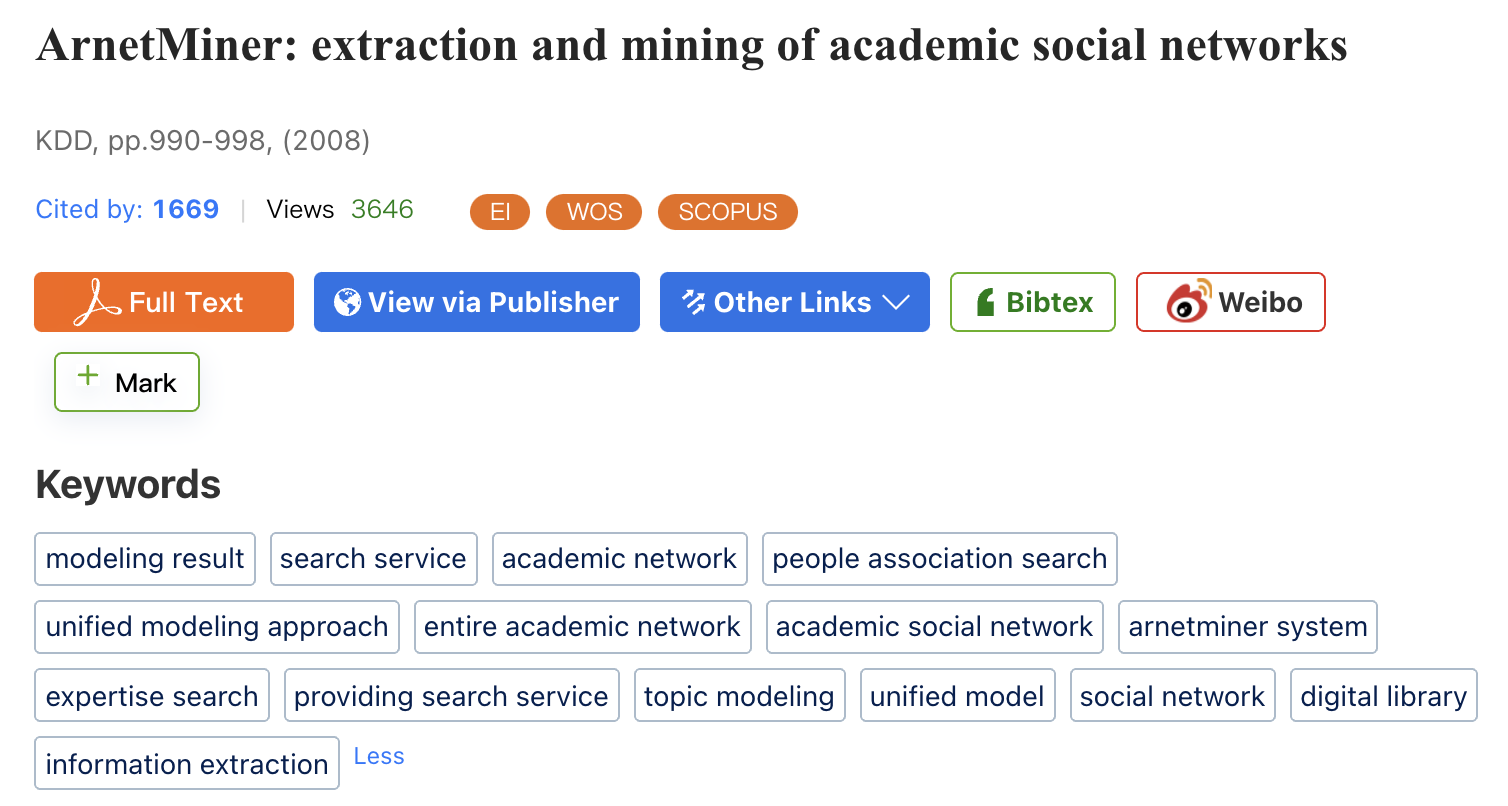}
    \caption{Deployed zero-shot paper tagging service in AMiner. \textmd{\method yields fine-grained tags at various lengths.}}
    \label{fig:paper_tagging}
\end{figure}

\vpara{Fields-of-study Tagging.}
Fields-of-study tagging, referred to as fields-of-study (FOS) linking, is a fundamental mission to associate unstructured scientific contents with structured disciplinary taxonomy (a case is presented in Figure~\ref{fig:paper_tagging}). Its results also serve as indispensable features for various downstream supervised applications. 

However, it is a notoriously arduous undertaking to discover new FOS from enormous corpora; in addition, how to continuously drive algorithms to link a paper with newly discovered FOS also remains largely unexplored. However, thanks to the massive entity knowledge \method has grasped in pre-training, it can be solved now using \method's zero-shot inference without a lift of fingers.

We present a case study, where \method is applied to tag the paper of GPT-3~\cite{brown2020language} given its title and abstract. Using beam search with a width of 16 to decode FOS entities, we search from single-token entities to quadruple-token entities. The top 16 generated ones are listed in Table~\ref{tab:zero-shot-case-study}. The groud truth (or namely \textit{gold}) FOS later annotated in MAG are all included in the top 16. Surprisingly, some fine-grained correct entities, though not in existing FOS, are also generated, such as \textit{Autoregressive language model} or \textit{Few shot learning}. Despite some ill-formed or inappropriate entities such as \textit{Architecture} or \textit{Artificial language processing}, \method's zero-shot tagging capability is still quite amazing.

To quantitatively evaluate the performance paper tagging, we adapt FOS prediction task from MAG. First, we choose 19 top-level field-of-studies (FOS) such as ``biology'' and ``computer science''. Then, from the paper data which were not used in the pre-training process, we randomly select 1,000 papers for each FOS. The task is to rank all FOS for each paper by estimating the probabilities of Equation~\ref{eq:entity-prob-decode} given paper title and optional abstract. 

We also apply two techniques to improve the model decoding performance. The first technique is to add extra \textit{prompt} word to the end of the paper title (before masked tokens). We select ``Field of study:'' as the prompt words in the FOS inference task. The second technique is to concatenate the paper abstract to the end of the paper title. We report the Hit@1 and MRR scores in Table~\ref{tab:zero-shot-result}.



\begin{table}[t]
  \centering
  \vcaption{\method's zero-shot paper tagging on the paper of GPT-3 given its title and abstract. \textmd{The groundtruth FOS are \textbf{bolded}. Newly created FOS by \method are \underline{underlined}.}}
  \label{tab:zero-shot-case-study}
  \footnotesize
  \begin{tabular}{@{}p{0.06\textwidth}p{0.38\textwidth}@{}}
  \toprule[1.2pt]
  Title & Language Models are Few-Shot Learners \\
  \midrule
  Abstract & Recent work has demonstrated substantial gains on many NLP tasks and benchmarks by pre-training on a large corpus of text followed by fine-tuning on a specific task. While typically task-agnostic in architecture, this method still requires task-specific fine-tuning datasets of thousands or tens of thousands of examples. By contrast, humans can generally... \\
  \midrule
  Generated FOS & Natural language processing, \underline{Autoregressive language model}, \textbf{Computer science}, \underline{Sentence}, Artificial intelligence, Domain adaptation, \textbf{Language model}, \underline{Few shot learning}, \underline{Large corpus}, Arithmetic, Machine learning, Architecture, Theoretical computer science, Data mining, \textbf{Linguistics}, \underline{Artificial language processing} \\
  \midrule
  Gold FOS & Language model, Computer science, Linguistics \\
  \bottomrule[1.2pt]
  \end{tabular}
\end{table}

\vpara{Venue and Affiliation Prediction.}
Analogously to paper tagging, venue and affiliation prediction of certain papers can also be conducted in zero-shot learning setting. From non-pretrained papers, we choose the 30 most frequent arXiv categories and 30 affiliations as inference candidates, with 100 papers randomly selected for each candidate. Full lists of the candidates including FOS candidates are enclosed in the appendix.

The experiment settings completely follow the FOS inference task, except that we use ``Journal or Venue:'' and ``Affiliations:'' as prompt words respectively. The entity type embeddings for masked entities in \method are also replaced by venue and affiliation entity type embeddings accordingly. 

In Table~\ref{tab:zero-shot-result}, we can see that the proposed augmented \method outperforms SciBERT by a large margin. Although SciBERT was not pre-trained with entity knowledge, it still performs much greater than a random guess, which means the inference tasks are not independent of the paper content information. We speculate that the pre-training process on paper content (as used in SciBERT) also helps the model learn some generalized knowledge on other types of information, such as field-of-studies or venue names.

We also observe that the proposed use of abstract can always help improve the performance. On the other hand, the prompt words works well with SciBERT but only provide limited help for \method. Besides, the affiliation inference task appears to be harder than the other two tasks. Further analysis are provided in the \ref{app:zero-shot}. Two extended experiments are enclosed as well,  which reveal two findings:
\begin{enumerate}[leftmargin=0.2in]
    \item Using the summation of token log probabilities as the entity log probability is better than using the average.
    \item The out-of-order decoding is more suitable for encoder-based models like SciBERT and \method, compared with the left-to-right decoding.
\end{enumerate} 

\begin{figure}[t]
    \centering
    \includegraphics[width=1\linewidth]{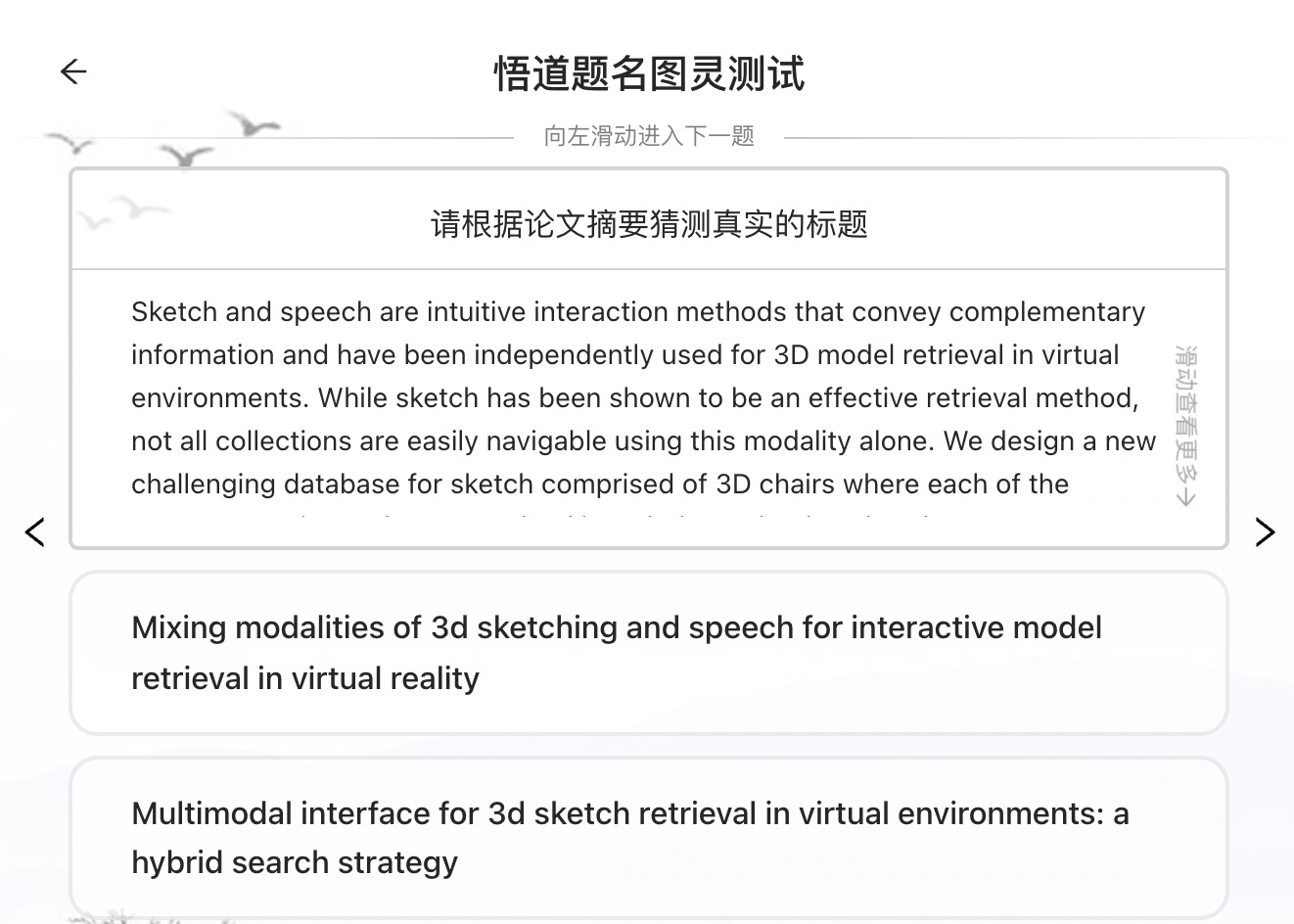}
    \caption{Deployed zero-shot Title Generation application for Turing Test in Wudao.$^7$ }
    \label{fig:title_generation}
    \vspace{-5mm}
\end{figure}

\vpara{Automatic Title Generation.}
How to summarize the contributions of a research paper into one sentence? Given the abstract, even senior experts may not figure out in a few seconds, but \method can generate titles comparable to original human-written ones in a zero-shot manner. During the pre-training,the span masking strategy will also be applied to titles, allowing \method to learn to summarize. Some case studies are presented in Table~\ref{tab:title_generation}, in which we observe that \method can generate quite the same title as origin given the paper abstract, even for our paper itself.

We also provide an interactive testing demo application online (as shown in Figure~\ref{fig:title_generation})\footnote{Try demo at: \url{https://wudao.aminer.cn/turing-test/v1/game/pubtitle}} to test if graduate-level students can distinguish between \method generated and original titles. Results suggest that there is probably only a  small gap in performance between \method's generation and human assignment's.


\begin{table}[t]
\caption{Upper: \textmd{case study in \method generated titles and original title. } Lower: \textmd{Online testing result from 660 random human views on 50 pairs of \method generated and original titles.}}
\footnotesize
\begin{tabular}{@{}lp{0.37\textwidth}<{\raggedright}@{}}
\toprule[1.2pt]
                         & \method Generated v.s. Original                                                               \\ \midrule
\multirow{2}{*}{\method} & OAG-LM: A Unified Backbone for Academic Knowledge Services                            \\
                         & OAG-LM: A Unified Backbone Language Model for Academic Knowledge Services             \\ \midrule
\multirow{2}{*}{AMiner}  & ArnetMiner: A System for Extracting and Mining Academic Social Networks              \\
                         & ArnetMiner: Extraction and Mining of Academic Social Networks                         \\ \midrule
\multirow{2}{*}{ResNet}  & Deep Residual Networks for Visual Recognition : A Comparison of Deep and VGG Networks \\
                         & Deep Residual Networks for Image Recognition                                          \\ \midrule
\multirow{2}{*}{SciBERT} & SciBERT: A Pretrained Language Model for Scientific NLP                              \\
                         & SciBERT: A Pretrained Language Model for Scientific Text                              \\ \bottomrule[1.2pt]
\end{tabular}
\vspace{1mm}

\normalsize
\small
\renewcommand\tabcolsep{12.5pt}
\begin{tabular}{@{}lccc@{}}
\toprule[1.2pt]
Method             & Total & Select & Selection Rate \\ \midrule
\method Generated & 330   & 157    & 47.6\%          \\
Original          & 330   & 163    & 52.4\%          \\ \bottomrule[1.2pt]
\end{tabular}
\label{tab:title_generation}
\vspace{-3mm}
\end{table}

\hide{
\begin{table}[ht]
  \centering
  \vcaption{The generated FOS for the paper of GPT-3. The gold FOS are bolded. FOS not in the original OAG FOS candidate list are underlined.}
  \label{tab:zero-shot-case-study}
  \scriptsize
  \begin{tabular}{p{0.05\textwidth}p{0.38\textwidth}}
  \toprule
  Title & Language Models are Few-Shot Learners \\
  \midrule
  Abstract & Recent work has demonstrated substantial gains on many NLP tasks and benchmarks by pre-training on a large corpus of text followed by fine-tuning on a specific task. While typically task-agnostic in architecture, this method still requires task-specific fine-tuning datasets of thousands or tens of thousands of examples. By contrast, humans can generally... \\
  \midrule
  Generated FOS & Natural language processing, \underline{Autoregressive language model}, \textbf{Computer science}, \underline{Sentence}, Artificial intelligence, Domain adaptation, \textbf{Language model}, \underline{Few shot learning}, \underline{Large corpus}, Arithmetic, Machine learning, Architecture, Theoretical computer science, Data mining, \textbf{Linguistics}, \underline{Artificial language processing} \\
  \midrule
  Gold FOS & Language model, Computer science, Linguistics \\
  \bottomrule
  \end{tabular}
\end{table}
}

\hide{
In summary, although our proposed \method model is not born for decoding, it still exhibits the potential of generating high-quality entities in the zero-shot settings.

\subsection{Supervised Classification}

\begin{table}[t]
  \small
  \centering
  \vcaption{The results of the classification task.}
  \label{tab:supervised-classification-result}
  \begin{threeparttable}
  \begin{tabular}{lcccc}
  \toprule
  \multirow{2}{*}{Tasks} & \multicolumn{2}{c}{Freeze} & \multicolumn{2}{c}{Finetune} \\
  & SciBERT & \method & SciBERT & \method \\
  \midrule
  \multicolumn{3}{l}{FOS} \\
  \quad\textit{title only} & $33.25^{0.25}$ & $\mathbf{43.28}^{0.12}$ & $\mathbf{55.13}^{0.30}$ & $54.54^{0.29}$ \\
  \quad\textit{+author} & $30.15^{0.07}$ & $\mathbf{41.87}^{0.06}$ & $\mathbf{55.63}^{0.42}$ & $55.30^{0.43}$ \\
  \quad\textit{+venue} & $34.77^{0.17}$ & $\mathbf{46.99}^{0.15}$ & $63.18^{0.18}$ & $\mathbf{63.53}^{0.08}$ \\
  \quad\textit{+aff} & $32.83^{0.13}$ & $\mathbf{43.07}^{0.11}$ & $\mathbf{55.06}^{0.21}$ & $54.65^{0.38}$ \\
  \quad\textit{+all} & $32.83^{0.08}$ & $\mathbf{45.47}^{0.16}$ & $63.43^{0.15}$ & $\mathbf{64.22}^{0.38}$ \\
  \cmidrule{1-5}
  \multicolumn{3}{l}{Venue} \\
  \quad\textit{title only} & $24.62^{0.52}$ & $\mathbf{32.87}^{1.47}$ & $61.86^{0.32}$ & $\mathbf{63.03}^{0.46}$ \\
  \quad\textit{+author} & $21.21^{0.82}$ & $\mathbf{30.91}^{0.96}$ & $62.62^{0.34}$ & $\mathbf{63.46}^{0.48}$ \\
  \quad\textit{+aff} & $24.38^{0.49}$ & $\mathbf{32.32}^{1.36}$ & $62.13^{0.43}$ & $\mathbf{62.65}^{0.49}$ \\
  \quad\textit{+fos} & $40.49^{1.25}$ & $\mathbf{52.61}^{0.79}$ & $78.05^{0.14}$ & $\mathbf{78.47}^{0.25}$ \\
  \quad\textit{+all} & $39.92^{1.17}$ & $\mathbf{51.33}^{0.44}$ & $77.88^{0.16}$ & $\mathbf{78.34}^{0.62}$ \\
  \cmidrule{1-5}
  \multicolumn{3}{l}{Affiliation} \\
  \quad\textit{title only} & $13.88^{0.83}$ & $\mathbf{19.72}^{0.64}$ & $\mathbf{35.44}^{0.45}$ & $35.04^{0.61}$ \\
  \quad\textit{+author} & $20.65^{1.04}$ & $\mathbf{32.19}^{0.92}$ & $52.68^{0.18}$ & $\mathbf{53.33}^{0.43}$ \\
  \quad\textit{+venue} & $16.57^{0.60}$ & $\mathbf{25.23}^{0.72}$ & $43.13^{0.36}$ & $\mathbf{43.65}^{0.40}$ \\
  \quad\textit{+fos} & $17.39^{0.86}$ & $\mathbf{22.06}^{0.37}$ & $37.05^{0.80}$ & $\mathbf{37.60}^{0.51}$ \\
  \quad\textit{+all} & $24.02^{0.87}$ & $\mathbf{32.49}^{0.50}$ & $56.04^{0.95}$ & $\mathbf{57.63}^{0.49}$ \\
  \bottomrule
  \end{tabular}
  \end{threeparttable}
  \vspace{-0.2in}
\end{table}

In this section, we develop the supervised classification tasks on top of the datasets described above, which are enlarged by 10 times following the same generating process. The data in the zero-shot inference are kept as test sets. We construct validation sets to select the best models during fine-tuning, with the same size as the test sets. The rest data are used as training sets. The sizes of all datasets for all tasks are enclosed in the appendix.

In supervised classification tasks, we remove the masked tokens and feed the averaged output embeddings from the pre-training models to a single fully-connected layer. We apply softmax layer to make predictions at last. 
As for the inputs, to present the effectiveness of heterogeneous entity types, we not only use paper titles as inputs but also concatenate other entities. 
Besides, we also tested the model performance with and without the original pre-training model parameters frozen. We follow the standard configurations for fine-tuning BERT, which are enclosed in the appendix.

As shown in Table~\ref{tab:supervised-classification-result}, the \method outperforms SciBERT by a large margin when the parameters in pre-trained parts are frozen. When not frozen, for venue and affiliation prediction, \method surpasses SciBERT significantly. In FOS prediction, although \method under-performs SciBERT in some cases, the best performance for using all available entities in \method still beats the one reached by SciBERT. 

We also observe that different types of entities contribute to various tasks in dissimilar ways. For example, the use of author information is particularly helpful for affiliation prediction but not very useful in FOS prediction. On the other hand, the field of study (FOS) inputs, work pretty well in venue prediction but provide marginal improvements to affiliation prediction.

In conclusion, the proposed \method is effective in both zero-shot tasks and supervised tasks. The additionally learned heterogeneous entities can help the model reach better performance while dealing with multiple types of inputs.

\input{nlp_benchmark_results.tex}

\begin{table}[ht]
    \small
    \centering
    \vcaption{The result of link prediction tasks.}
    \label{tab:hgt}
    \begin{tabular}{lcccc}
    \toprule
    \multirow{2}{*}{Tasks} & \multicolumn{2}{c}{Paper-Field} & \multicolumn{2}{c}{Paper-Venue} \\
    & NDCG & MRR & NDCG & MRR \\
    \midrule
    XLNet & 0.3939 & 0.4473 & 0.4385 & 0.2584 \\
    SciBERT & 0.4740 & 0.5743 & 0.4570 & 0.2834 \\
    OAG-BERT & \textbf{0.4892} & \textbf{0.6099} & \textbf{0.4844} & \textbf{0.3131} \\
    \bottomrule
    \vspace{-0.2in}
    \end{tabular}
\end{table}

\subsection{Link Prediction}


In previous sections, we present the effectiveness of using OAG-BERT individually. In this section, we apply the heterogeneous entity embeddings of OAG-BERT as pre-trained initializations for node embeddings on the academic graph and show that OAG-BERT can also work together with other types of models. Specifically, we take the heterogeneous graph transformer (HGT) model from~\cite{hu2020heterogeneous} and combine it with the pre-trained embeddings from OAG-BERT.

To make predictions for the links in the heterogeneous graph, the authors of HGT first extract node features and then apply HGT layers to encode graph features. For paper nodes, the authors use XLNet~\cite{yang2019xlnet} to encode titles as input features. For other types of nodes, HGT use metapath2vec~\cite{dong2017metapath2vec} to initialize the features.

However, there are two problems with using XLNet on the heterogeneous academic graph. First, the XLNet was pre-trained on universal language corpus, which is lack of academic domain data. Second, XLNet can only encode paper nodes by using their titles and is unable to generate useful embeddings for other types of nodes like author or affiliation.

To this end, we propose to replace the original XLNet encoder with our OAG-BERT model, which can tackle the two challenges mentioned above. We use the OAG-BERT model to encode all types of nodes and use the generated embeddings as their node features. To prove the effectiveness of OAG-BERT on encoding heterogeneous nodes, we also compare the performance of SciBERT with OAG-BERT. We experimented on the CS dataset released by HGT\footnote{https://github.com/acbull/pyHGT}. The details of the dataset are delivered in the appendix.

The NDCG and MRR scores for the Paper-Field and Paper-Venue link prediction are reported in Table~\ref{tab:hgt}. It shows that SciBERT surpasses the original XLNet performance significantly, due to the pre-training on the large scientific corpus. Our proposed OAG-BERT made further improvements on top of that, as it can better understand the entity knowledge on the heterogeneous graph.

\subsection{NLP Tasks}

Previous experiments have demonstrated the superiority of OAG-BERT on tasks involving multi-type entities. In this section, we will further explore the performance of OAG-BERT on natural language processing tasks, which only contain text-based information such as paper titles and abstracts. We will show that although pre-trained with heterogeneous entities, the OAG-BERT can still perform competitive results with SciBERT on NLP tasks.


We made comparisons over three models, including \textbf{SciBERT} (both the original paper results and the reproduced results), \textbf{S2ORC} (similar to SciBERT except pre-trained with more data), and \textbf{OAG-BERT} (both the vanilla version and the augmented version).


In accord with SciBERT~\cite{beltagy2019scibert}, we evaluate the model performance on the same 12 NLP tasks, including Named Entity Recognition (NER), Dependency Parsing (DEP), Relation Extraction (REL), PICO Extraction (PICO), and Text Classification (CLS). These tasks only focus on single sentence representation so we add another three sequential sentence classification (SSC) tasks used in~\cite{cohan2019pretrained}, to further verify the capability of pre-training models on long texts. The evaluation metrics are also accord with the usage in SciBERT~\cite{beltagy2019scibert} and Sequential-Sentence-Classification~\cite{cohan2019pretrained}, which can be found in the appendix along with the task details and hyper-parameter settings.

The results in Table~\ref{tab:nlp-benchmarks} show that the proposed OAG-BERT is competitive with SciBERT and a bit behind the S2ORC. Comparing with the reproduced SciBERT, our vanilla OAG-BERT only shows clear disadvantages on the SciERC REL task and the ACL-ARC CLS task, where datasets are relatively small and are sensitive to a few swinging samples. We ascribe the minor differences in other tasks to the differences in training corpus and the data cleaning techniques. The augmented OAG-BERT, although trained with heterogeneous entities that differ from the inputs of downstream NLP tasks, still presents similar performance to the vanilla version.

In summary, despite the fact that the OAG-BERT does not surpass the previous state-of-the-art academic pre-training model on NLP tasks, it still keeps the knowledge on these language dedicated tasks even after pre-training with multiple types of entities.
}






\section{Deployed Applications}

In this section, we will introduce several real-world applications where our \method is employed.

First, the results on the name disambiguation tasks indicate that the \method is relatively strong at encoding paper information with multi-type entities, which further help produce representative embeddings for the paper authors. Thus, we apply the \method to the NSFC reviewer recommendation problem~\cite{cyranoski2019artificial}. The National Natural Science Foundation of China is one of the largest science foundations, where an enormous number of applications are reviewed every year. Finding appropriate reviewers for applications is time-consuming and laborious. To tackle this problem, we collaborate with Alibaba and develop a practical algorithm on top of the \method which can automatically assign proper reviewers to applications and greatly benefits the reviewing process.

In addition to that, we also integrate the \method as a fundamental component for the AMiner~\cite{tang2008arnetminer} system. In AMiner, we utilize \method to handle rich information on the academic heterogeneous graph. For example, with the ability of decoding FOS entities, we use the \method to automatically generate FOS candidates for unlabeled papers. Besides, we similarly amalgamate the \method into the name disambiguation framework. Finally, we employ \method to recommend related papers for users, leveraging its capability in encoding paper embeddings. The \method model is also released in CogDL package.

\section{Conclusion}

In conclusion, we propose \method, a heterogeneous entity-augmented language model to serve as the backbone for academic knowledge services. It incorporates entity knowledge during pre-training, which benefits many downstream tasks involving strong entity knowledge. \method is applied to nine typical  academic applications and also deployed in AMiner  and for NSFC grant reviewer recommendation. 
We release the pre-trained \method model in CogDL for free use to everyone. 


\section*{ACKNOWLEDGEMENT}
We thank the reviewers for their valuable feedback to improve this work. 
This work is supported by Technology and Innovation Major Project of the Ministry of Science and Technology of China under Grant 2020AAA0108400 and 2020AAA0108402, Natural Science
Foundation of China (Key Program, No. 61836013), and National Science
Foundation for Distinguished Young Scholars (No. 61825602).

\bibliographystyle{ACM-Reference-Format}
\balance
\bibliography{ref.bib}

\clearpage
\newpage
\appendix
\section{Experiment Supplementary}


\subsection{Zero-Shot Inference} \label{app:zero-shot}


\vpara{Use of Prompt Word}
As shown in Table~\ref{tab:zero-shot-result}, the use of proposed prompt words in the FOS inference task, turns out to be fairly useful for SciBERT to decode paper fields (FOS). We conjecture it is because the extra appended prompt words can help alter the focus of the pre-training model while making predictions on masked tokens. However, the improvement for SciBERT is marginal on affiliation inference. When decoding venue, it even hurts the performance. This is probably due to the improper choice of prompt words.

For OAG-BERT, this technique has limited help as our expectation. Instead of using continuous positions as SciBERT, OAG-BERT encodes inter-entity positions to distinguish different entities and paper texts. Thus the additional appended prompt word is treated as part of the paper title and is not adjacent to the masked entities for OAG-BERT.


\vpara{Use of Abstract}
The use of abstracts can greatly improve the model inference performance in both SciBERT and OAG-BERT. Both models frequently accept long text inputs in the pre-training process, which makes them naturally favor abstracts. Besides, abstracts contain rich text information which can help the pre-training model capture the main idea of the whole paper.

\vpara{Task Comparisons}
The affiliation generation task appears to be much harder than the other two tasks. This is probably due to the weak semantic information contained in affiliation names. The words in field-of-studies can be seen as sharing the same language with paper contents and most venue names also contain informative concept words such as ``Machine Learning'' or ``High Energy''. This is not always true for affiliation names. For universities like ``Harvard University'' or ``University of Oxford'', their researchers could focus on multiple unrelated domains which are hard for language models to capture. For companies and research institutes, some may focus on a single domain but it is not necessary to have such descriptions in their names, which also confuses the pre-training language model.

\vpara{Discussion for Entity Probability}
In Equation~\ref{eq:entity-prob-decode}, we use the sum of log probabilities of all tokens to calculate the entity log probability. This method seems unfair for entities with longer lengths as the log probability for each token is always negative. However, for MLM-based models, the encoding process not only encodes ``[MASK]'' tokens but also captures the length of the masked entity and each token's position. Therefore, if the pre-training corpus has fewer long entities than short entities, in the decoding process, the decoded tokens in a long entity will generally receive higher probability, compared to the ones in a short entity.

Even so, the sum of log probabilities is still not necessary to be the best choice depending on the entity distribution in the pre-training corpus. We conduct a simple experiment to test different average methods. We reform the calculation of entity log probability in Equation~\ref{eq:entity-prob-decode} as 
$
  \frac{1}{L^\alpha}\sum_{1\le k\le l}\log P(w_{i_k}|\mathcal{C},w_{i_1},w_{i_2},...,w_{i_{k-1}})
$
, where $L$ denotes the length of the target entity. When $\alpha=0$, this equation degrades to the summation version used in previous tasks. When $\alpha=1$, this equation degrades to the average version.

We compare different averaging methods by using various $\alpha$
and test their performance on the zero-shot inference tasks. We select the input features with the best performance according to Table~\ref{tab:zero-shot-result}. For SciBERT, we use both abstract and prompt word for FOS and affiliation inference. We do not use the prompt word for venue inference. For OAG-BERT, we only use abstract as the prompt word does not work well. The results in Table~\ref{tab:avg-or-sum-logprob} show that for the most time, using the summation strategy outperforms the average strategy significantly. The simple average ($\alpha=1$) appears to be the worst choice. However, for some situations, a moderate average ($\alpha=0.5$) might be beneficial.

\begin{table}[ht]
  \centering
  \vcaption{The results for using different average methods while calculating entity log probabilities. Hit@1 and MRR are reported.}
  \small
  \begin{tabular}{lcccc}
  \toprule
  Method & $\alpha=0$ & $\alpha=0.5$ & $\alpha=1$ \\
  \midrule
  SciBERT \\
  \textit{FOS} & \textbf{35.33\%, 0.52} & 32.07\%, 0.51 & 14.85\%, 0.36 \\
  \textit{Venue} & 18.00\%, 0.32 & \textbf{19.30\%, 0.33} & 7.07\%, 0.23 \\
  \textit{Affiliation} & \textbf{12.40\%, 0.25} & 10.83\%, 0.23 & 9.23\%, 0.21 \\
  \midrule
  OAGBERT \\
  \textit{FOS} & \textbf{49.59\%, 0.67} & 48.08\%, 0.66 & 45.36\%, 0.63 \\
  \textit{Venue} & \textbf{39.00\%, 0.57} & 38.20\%, 0.57 & 36.13\%, 0.55 \\
  \textit{Affiliation} & \textbf{21.67\%, 0.38} & 19.90\%, 0.36 & 16.47\%, 0.31 \\
  \bottomrule
  \end{tabular}
  \label{tab:avg-or-sum-logprob}
\end{table}


\vpara{Discussion for Decoding Order}
In our designed decoding process, we do not strictly follow the left-to-right order as used in classical decoder models. The main reason is that for encoder-based BERT model, the decoding for each masked token relies on all bidirectional context information, rather than only prior words. We compare the performance of using left-to-right decoding and out-of-order decoding in Table~\ref{tab:forward-or-not-full}.

The results show that for FOS, there is no significant difference between two decoding orders, since the candidate FOS only has one or two tokens inside. As for venue and affiliation, it turns out that the out-of-order decoding generally performs much better than left-to-right decoding, except when OAG-BERT uses abstract where differences are relatively small as well. We also present the results for models using left-to-right decoding and prompt words in Table~\ref{tab:forward-or-not-full}, which indicates that the left-to-right decoding will sometimes undermine the effectiveness of prompt words significantly, especially for OAG-BERT.

\FloatBarrier
\begin{table}[h]
    \small
    \centering
    \vcaption{The results for using left-to-right decoding and out-of-order decoding order. Hit@1 and MRR are reported. Results with difference larger than 1\% Hit@1 were bolded.}
    \label{tab:forward-or-not-full}
    \begin{tabular}{lcccccc}
    \toprule
    Method & \multicolumn{2}{c}{FOS} & \multicolumn{2}{c}{Venue} & \multicolumn{2}{c}{Affiliation} \\
    & Hit@1 & MRR & Hit@1 & MRR & Hit@1 & MRR \\
    \midrule
    \multicolumn{3}{l}{SciBERT}\\
    \quad \textit{Left-to-Right} & 20.05\% & 0.37 & 8.40\% & 0.20 & 6.90\% & 0.18 \\
    \quad \textit{Out-of-Order} & 19.93\% & 0.37 & \textbf{9.87\%} & \textbf{0.22} & 6.93\% & 0.19 \\
    \midrule
    \multicolumn{3}{l}{SciBERT \textit{+prompt}}\\
    \quad \textit{Left-to-Right} & 29.65\% & 0.47 & 9.57\% & 0.21 & 8.03\% & 0.20 \\
    \quad \textit{Out-of-Order} & 29.59\% & 0.47 & 10.03\% & 0.21 & 8.00\% & 0.20 \\
    \midrule
    \multicolumn{3}{l}{SciBERT \textit{+abstract}}\\
    \quad \textit{Left-to-Right} & 25.67\% & 0.43 & 11.43\% & 0.24 & 7.63\% & 0.19 \\
    \quad \textit{Out-of-Order} & 25.66\% & 0.43 & \textbf{18.00\%} & \textbf{0.32} & \textbf{10.33\%} & \textbf{0.22} \\
    \midrule
    \multicolumn{3}{l}{SciBERT \textit{+both}}\\
    \quad \textit{Left-to-Right} & 35.21\% & 0.52 & \textbf{11.17\%} & \textbf{0.24} & 11.47\% & 0.23 \\
    \quad \textit{Out-of-Order} & 35.33\% & 0.52 & 9.83\% & 0.22 & 12.40\% & 0.25 \\
    \midrule
    \multicolumn{3}{l}{OAG-BERT}\\
    \quad \textit{Left-to-Right} & 34.94\% & 0.53 & 11.33\% & 0.24 & 5.47\% & 0.17 \\
    \quad \textit{Out-of-Order} & 34.36\% & 0.51 & \textbf{21.00\%} & \textbf{0.37} & \textbf{11.03\%}& \textbf{0.24} \\
    \midrule
    \multicolumn{3}{l}{OAG-BERT \textit{+prompt}}\\
    \quad \textit{Left-to-Right} & 37.84\% & 0.56 & 12.53\% & 0.26 & 5.50\% & 0.17 \\
    \quad \textit{Out-of-Order} & 37.33\% & 0.55 & \textbf{22.67\%} & \textbf{0.39}, & \textbf{11.77\%} & \textbf{0.25}\\
    \midrule
    \multicolumn{3}{l}{OAG-BERT \textit{+abstract}}\\
    \quad \textit{Left-to-Right} & 49.75\% & 0.67 & \textbf{40.50\%} & \textbf{0.59} & 21.93\% & 0.38 \\
    \quad \textit{Out-of-Order} & 49.59\% & 0.67 & 39.00\% & 0.57 & 21.67\% & 0.38 \\
    \midrule
    \multicolumn{3}{l}{OAG-BERT \textit{+both}}\\
    \quad \textit{Left-to-Right} & 49.83\% & 0.67 & 22.17\% & 0.38 & 6.80\% & 0.19 \\
    \quad \textit{Out-of-Order} & 49.51\% & 0.67 & \textbf{38.47\%} & \textbf{0.57} & \textbf{21.53\%} & \textbf{0.38} \\
    \bottomrule
    \end{tabular}
\end{table}
\FloatBarrier

\FloatBarrier
\begin{table}[H]
    \small
    \centering
    \vcaption{A full list of used candidates in zero-shot inference tasks and supervised classification tasks.}
    \label{tab:zero-shot-candidates}
    \begin{tabular}{p{0.45\textwidth}}
    \toprule
    \textbf{FOS}: Art, Biology, Business, Chemistry, Computer science, Economics, Engineering, Environmental science, Geography, Geology, History, Materials science, Mathematics, Medicine, Philosophy, Physics, Political science, Psychology, Sociology \\
    \midrule
    \textbf{Venue}: Arxiv: algebraic geometry, Arxiv: analysis of pdes, Arxiv: astrophysics, Arxiv: classical analysis and odes, Arxiv: combinatorics, Arxiv: computer vision and pattern recognition, Arxiv: differential geometry, Arxiv: dynamical systems, Arxiv: functional analysis, Arxiv: general physics, Arxiv: general relativity and quantum cosmology, Arxiv: geometric topology, Arxiv: group theory, Arxiv: high energy physics - experiment, Arxiv: high energy physics - phenomenology, Arxiv: high energy physics - theory, Arxiv: learning, Arxiv: materials science, Arxiv: mathematical physics, Arxiv: mesoscale and nanoscale physics, Arxiv: nuclear theory, Arxiv: number theory, Arxiv: numerical analysis, Arxiv: optimization and control, Arxiv: probability, Arxiv: quantum physics, Arxiv: representation theory, Arxiv: rings and algebras, Arxiv: statistical mechanics, Arxiv: strongly correlated electrons \\
    \midrule
    \textbf{Affiliation}: Al azhar university, Bell labs, Carnegie mellon university, Centers for disease control and prevention, Chinese academy of sciences, Electric power research institute, Fudan university, Gunadarma university, Harvard university, Ibm, Intel, Islamic azad university, Katholieke universiteit leuven, Ludwig maximilian university of munich, Max planck society, Mayo clinic, Moscow state university, National scientific and technical research council, Peking university, Renmin university of china, Russian academy of sciences, Siemens, Stanford university, Sun yat sen university, Tohoku university, Tsinghua university, University of california berkeley, University of cambridge, University of oxford, University of paris \\
    \bottomrule
    \end{tabular}
\end{table}
\FloatBarrier

\FloatBarrier
\begin{table}[H]
    \small
    \vcaption{The sizes for datasets used in supervised classification tasks.}
    \label{tab:classification-dataset-size}
    \centering
    \begin{tabular}{lllll}
    \toprule
    Task & Categories & Train & Validation & Test \\
    \midrule
    FOS & 19 & 152000 & 19000 & 19000 \\
    Venue & 30 & 24000 & 3000 & 3000 \\
    Affiliation & 30 & 24000 & 3000 & 3000 \\
    \bottomrule
    \end{tabular}
\end{table}
\FloatBarrier

\FloatBarrier
\begin{table}[h]
    \small
    \vcaption{Details for the CS heterogeneous graph used in the link prediction.}
    \label{tab:cs-hgt-graph-size}
    \centering
    \begin{tabular}{llll}
    \toprule
     & Papers & Authors & FOS \\
    Nodes & 544244 & 510189 & 45717 \\
    \cmidrule{2-4}
    1116163 & Venues & Affiliations & \\
     & 6934 & 9079 & \\
    \midrule
     & \#Paper-Author & \#Paper-FOS & \#Paper-Venue \\
    \#Edges & 1862305 & 2406363 & 551960 \\
    \cmidrule{2-4}
    6389083 & \#Author-Affiliation & \#Paper-Paper & \#FOS-FOS \\
    & 519268 & 992763 & 56424 \\
    \bottomrule
    \end{tabular}
    \vspace{-0.1in}
\end{table}
\FloatBarrier

\end{document}